\documentclass[letterpaper, 10 pt, conference]{ieeeconf}

\IEEEoverridecommandlockouts                    
\usepackage{amsmath,amsfonts} % Math packages
\usepackage{siunitx}
\usepackage{subfig}
\usepackage{float} 
\usepackage{threeparttable}
\usepackage{graphicx}         \usepackage{xcolor}
\usepackage{multirow}
\usepackage{array}    
\usepackage{algorithmic}
\usepackage[linesnumbered, ruled]{algorithm2e}
\usepackage[backend=bibtex,style=ieee,natbib=true,sorting=none,url=false,doi=false,isbn=false]{biblatex} 
\AtBeginBibliography{\small}
\title{\LARGE \bf
Mechanical features based object recognition %with a robot
}

\author{Pakorn Uttayopas, Xiaoxiao Cheng, Jonathan Eden, and Etienne Burdet
\thanks{This work was supported in part by the EC grants FETOPEN 829186 PH-CODING and ITN PEOPLE 861166 INTUITIVE.}
\thanks{The authors are with the Department of Bioengineering, Imperial College of Science, Technology and Medicine, SW72AZ London, UK. Email: \{pu18, xcheng4, j.eden, e.burdet\}@imperial.ac.uk)}
}

\begin{document}

\maketitle
\thispagestyle{empty}
\pagestyle{empty}

%%%%%%%%%%%%%%%%%%%%%%%%%%%%%%%%%%%%%%%%%%%%%%%%%%%%%%%%%%%%%%%%%%%%%%%%%%%%%%%%
\begin{abstract}
Current robotic haptic object recognition relies on statistical measures derived from movement dependent interaction signals such as force, vibration or position. Mechanical properties that can be identified from these signals are intrinsic object properties that may yield a more robust object representation. Therefore, this paper proposes an object recognition framework using multiple representative mechanical properties: the coefficient of restitution, stiffness, viscosity and friction coefficient. These mechanical properties are identified in real-time using a dual Kalman filter, then used to classify objects. The proposed framework was tested with a robot identifying 20 objects through haptic exploration. The results demonstrate the technique's effectiveness and efficiency, and that all four mechanical properties are required for best recognition yielding a rate of 98.18\,$\pm$\,0.424\,\%. Clustering with Gaussian mixture models further shows that using these mechanical properties results in superior recognition as compared to using statistical parameters of the interaction signals.
\end{abstract}

\vspace{0.2cm}
\noindent \textbf{Keywords}: haptic exploration, interaction mechanics, feature extraction, supervised learning for classification, clustering 

%%%%%%%%%%%%%%%%%%%%%%%%%%%%%%%%%%%%%%%%%%%%%%%%%%%%%%%%%%%%%%%%%%%%%%%%%%%%%%%%
\section{Introduction}
As robots are increasingly used in various fields such as agriculture, they have to manipulate objects of different mechanical properties skillfully. For instance, to harvest tomatoes or potatoes with similar shape, it is necessary to know their respective mechanical properties to handle them without dropping or crushing them. To recognize the objects a robot is interacting with, it is necessary to extract the unique features that characterize them \cite{Luo2017Mechatronics}. While geometric features can be used to identify solid objects \cite{Allen1989, Martinez-Hernandez2018, Pezzementi2011}, the shape of compliant objects changes with interaction such that shape alone is not sufficient to identify them. 

It is therefore necessary to characterize objects through mechanical parameters extracted during interaction. Compliant objects can be recognised by using tactile information obtained during haptic interaction such as force and vibrations. Empirical measures of these signals, such as the maximum, minimum and variance have been used for classification \cite{Hoelscher2015, Dallaire2014, Kaboli2019}. While these interaction features can be used for object recognition, their value depends on specific actions, and their use can be highly redundant leading to high computational cost.

The intrinsic mechanical features of objects may yield a more specific representation and thus lead to a more efficient recognition. These material properties describe an object's behavior in response to a load, for example the energy loss during impact can be characterized by the coefficient of restitution \cite{Stronge2018}. The deformation and restoration of the surface in response to a force exerted perpendicular is characterised by the material's viscoelasticity. Similarly when applying a tangential force, the resistance to sliding can be characterized through the roughness. These parameters have been previously considered to estimate mechanical properties from tactile information.

The coefficient of restitution is an important property in characterising how a body reacts during impact. While this property has rarely been used for object recognition, related features have been extracted from acoustic and acceleration data by investigating signal magnitude in the frequency domain \cite{Rebguns2011, Neumann2018} or applying unsupervised learning methods \cite{Luo2017Neurocomputing} or statistical tools \cite{Roy1996, Wisanuvej2014}. The consideration of acceleration peak has also been used as a similar impact related measure for object recognition, proving able to recognise five different materials\cite{Windau2010}.

Compliance-related features characterise deformation in response to continuous forces. Empirically, these features can be estimated by analyzing the normal force signal during interaction \cite{Xu2013, Spiers2016, Kaboli2019}. Such approaches have been used to estimate stiffness \cite{Su2012, Bednarek2021} and to infer how full a bottle is when grasped \cite{Chitta2011}. Stiffness, however, only characterises the static response. To estimate both stiffness and viscosity, a recursive least-square algorithm has been used \cite{Love1995, Ren2017, Rossi2016} as well as a Gaussian process \cite{Chalasani2018}. This estimation has then been applied to object recognition in simulation \cite{Yane2022}.  

To characterise the response to sliding roughness related features are typically used.  These features are typically estimated through the force or vibration occurring in the tangential direction during sliding \cite{Huang2021, Chu2015, Sinapov2011}. In addition, a constant Coulomb model is commonly used to identify a surface's friction \cite{Sundaralingam2021, Su2015}. Using the surface's friction along with geometrical information, a robot could recognise 18 household objects with different shapes and materials \cite{Sun2018}. By considering dynamic friction parameters, the robot-environment interaction can be modelled as quasi-static LuGe model \cite{CanudasdeWit1995}. Using such dynamic friction parameters can benefit the classification of objects with different surface's materials \cite{Song2012, Liu2012IROS}.

These previous works exhibit how a single mechanical property can be used to recognise objects. However, multiple objects may exhibit the same value of a specific mechanical property and thus cannot be distinguished by it. Integrating the collection of mechanical property estimations into a haptic exploration framework may improve object recognition. However, there is currently no method to estimate multiple mechanical properties simultaneously and use them together. Furthermore, the coefficient of restitution has rarely been used to recognise objects.

This prompted us to develop a framework for the identification of mechanical properties and their use to recognise specific objects, which will be presented in this paper. In this new approach, the coefficient of restitution, stiffness, viscosity and friction coefficient are estimated from the interaction force during haptic exploration. Our work builds upon \cite{Li2018} which adapted viscoelastic parameters to maintain a stable interaction.  To address issues with parameter oscillation we used a dual Kalman filter to consider sensory noise. We further added estimation of the coefficients of friction and restitution. The resulting method is first validated in simulation. The role of each mechanical parameters in object recognition is then investigated before our method is compared to representative statistical and empirical methods of the literature.

\begin{figure}[tb]
\centering
\qquad \qquad
\includegraphics[width=1\columnwidth]{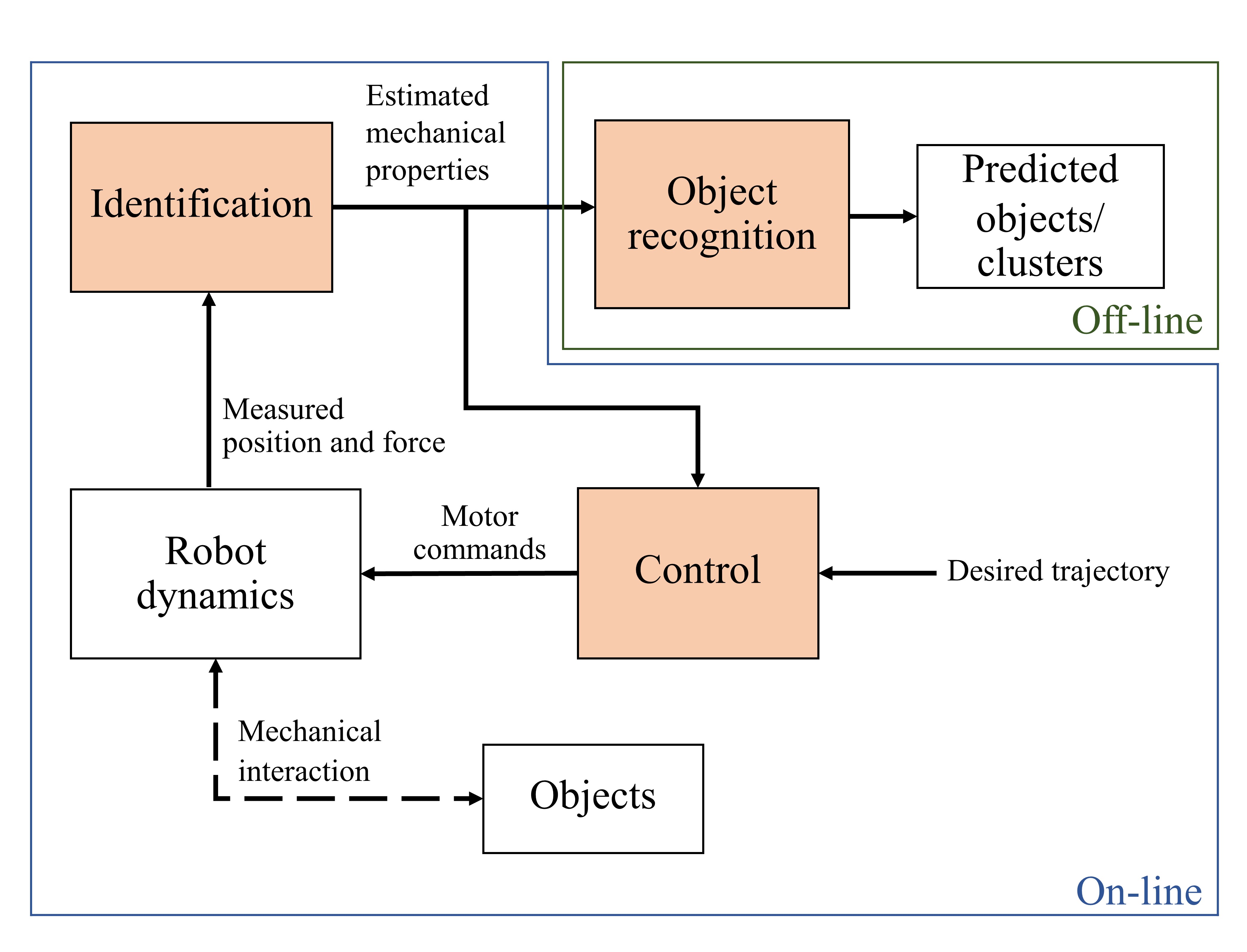}
\caption{Diagram of the object recognition process. The end-effector force and position measured during interaction are utilized to identify mechanical properties using an estimator. The estimated mechanical features are then used to recognize objects and adjust the motor command with the controller.}
\label{fig:diagram}
\end{figure}

%\section{Object recognition framework}
Fig.\,\ref{fig:diagram} shows the overall recognition framework with its  three components; identification, control, and object recognition. These components work together to identify and classify unknown objects based on their mechanical properties.

A robot, driven by the controller, interacts mechanically with objects to retrieve interaction force data as well as its positions. The robot's estimator first estimates the coefficient of restitution when touching the object's surface. A dual extended Kalman filter (DEKF) is then used to identify the object's stiffness, viscosity and friction coefficient online from signals of haptic sensors. These mechanical features are also used to adapt the controller's parameters so as to interact with each object properly. 

Additionally, the features based on the estimated mechanical parameters are combined to form the dataset feeding object recognition algorithms, in order to identify and cluster objects. This process is performed offline after the haptic exploration. 

% figure

%%%%%%%%%%%%%%%%%%%%%%%%%%%%%%%%%
\section{Online estimation and control}
\label{sec:estimator}
This section describes the online estimation and control. First, a discrete impact model and a continuous interaction model are introduced to capture the robot-environment interaction at different stages. Using these models, the estimation of impact (coefficient of restitution) and continuous interaction properties (stiffness, viscosity and friction coefficient) are presented. Finally, the interaction controller used to drive the robot interacting with the environment smoothly is explained.

\subsection{Interaction model}
Let the dynamics of a $n$-DOF robot interacting with its environment be described by
\begin{equation}
M(q)\ddot{x} + C(q,\dot{q})\dot{x} + G(q) \,=\, u + F + \omega\,,
\label{eq:robot_dynamics}
\end{equation}
where $x$ is the coordinate of the end effector in operational space and $q$ is the vector of joint angle. \(M(q)\) and \(C(q,\dot{q})\)  represent the inertia and Coriolis matrices and \(G(q)\) the gravitation vector, \(u\) is the control input and \(\omega\) motor noise. The interaction force \(F\) can be modelled with a mass-spring-damper system in the normal direction and Coulomb friction in the tangential direction: 
\begin{equation}
\begin{split}
F = \begin{bmatrix}
F_\perp\\
F_\parallel
\end{bmatrix}
& = \begin{bmatrix}
F_0 + \kappa\,x + d\,\dot{x}\\
\mu F_\perp
\end{bmatrix}, 
\end{split}
\label{eq:interaction_force}
\end{equation}
where $F_0 = - \kappa\,x_0$ is the force corresponding to the surface rest length $x_0$ (without interaction), $\kappa$ is the surface stiffness, \(d\) its viscosity, and \(\mu\) its friction coefficient.

\subsection{Impact estimation}
The initial contact of a robot with an object occurs in two phases: deformation and restoration. The deformation phase occurs before the initial point of contact and continues until maximum deformation. It is followed by the restoration from the time of maximum deformation until when separation occurs. By investigating impulses of these two phases, the coefficient of restitution is defined as a ratio of a normal impulse of restoration to a normal impulse of deformation \cite{Poisson1817}:
\begin{equation}
\widehat{\psi} = \, \frac{R}{D} \,=\, \left|\frac{\int_{t^0}^{t^+} \!\! F_\perp \,dt}{m_\perp [ \dot{x}_{\perp}(t^0) + \dot{x}_{\perp}(t^-)]}\right| \,.
\label{eq:CoR} 
\end{equation}
Here, $D$ is the momentum from 0.01\,s before collision, $t^-$, to the time of maximum deformation, $t^0$. $R$ integrates the normal force from $t^0$ to 0.01\,s after collision, $t^+$.

\subsection{Continuous properties estimation}
We assume that the robot can measure the end-effector position (e.g. from joint encoders) as well as the force normal to the surface subjected to a large noise $\nu$.

For the estimation, the system dynamics become nonlinear due to the coupling of the robot's states and mechanical parameters in the interaction force model (\ref{eq:interaction_force}). In discrete state-space  form, the dynamics of the robot interacting with the environment is:
\begin{eqnarray}
\label{eq:state_space_aug}
\xi_{k+1} \!\!\!\!&=&\!\!\!\! f(\xi_k,u_k,\theta_k) + \omega_k \nonumber \\
\eta_k \!\!\!\!&=&\!\!\!\! h(\xi_k) + \nu_k \\
\xi \!\!\!\!& \equiv &\!\!\!\! \left[\!\! \begin{array}{c} x_\perp \\ \dot{x}_\perp \\ x_\parallel \\ \dot{x}_\parallel \\ \mu \end{array} \!\!\right], \,\,\,
\eta \equiv \left[ \!\! \begin{array}{c} x_\perp \\ x_\parallel \end{array} \!\! \right], \,\,\,
u \equiv \left[ \!\! \begin{array}{c} u_\perp \\ u_\parallel \end{array} \!\! \right], \,\,\,
\theta \equiv \left[ \!\! \begin{array}{c} F_0 \\ k \\ d \end{array} \!\! \right] \nonumber
\end{eqnarray}
where $f$ is a nonlinear mapping obtained from (\ref{eq:robot_dynamics}) and $h$ is a nonlinear mapping between the states and observation. The augmented state $\xi$ consists of the robot's states and the friction parameters, $u$ are motor commands, where $\eta$ is the measured robot's positions and $\theta$ is the viscoelasticity vector.

Due to the system's nonlinearity, noise and the coupling between the states and parameters, we employ the dual extended Kalman filter method \cite{Wan2001} to estimate the robot's state and interaction mechanics' parameters simultaneously. The dual Kalman filter is a recursive estimation process, which uses partial measurements to estimate the parameters in the model before integrating the updated model and measurements to estimate the hidden states.

\begin{figure}[tb]
\centering
\qquad \qquad
\includegraphics[width=1\columnwidth]{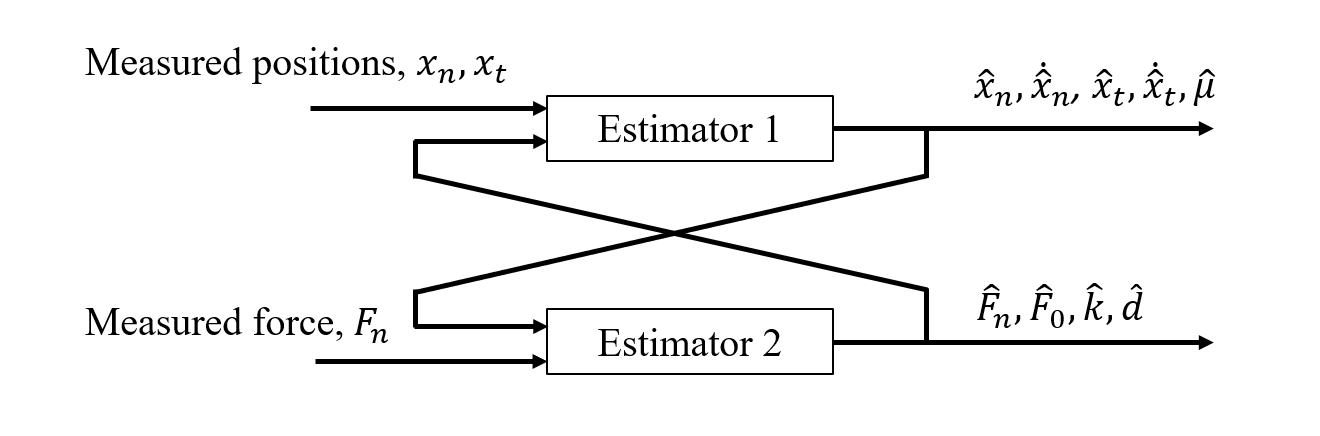}
\caption{Diagram of dual extended Kalman filter combining two estimators. At every time step, Estimator 1 identifies the robot states and friction coefficient based on the measured position and estimated interaction force. Estimator 2 identifies the object elastic-viscosity parameters and interaction force based on the measured force and estimated robot state.}
\label{fig:diagram_2}
\end{figure}

Fig.\,\ref{fig:diagram_2} depicts the two designed estimators. Estimator 1 estimates the state  $\xi$, which includes the robot's states and friction parameter. Estimator 2 then estimates the viscoelasticity parameters $\theta$ from the measured normal force. In principle, the prediction error cost will be minimized when the estimated parameters $\hat{\theta}, \hat{\mu}$ converge on the real values $\theta, \mu$ while the estimated states $\hat{\xi}$ converge to the real states $\xi$.  
% figure

%figure
\begin{figure*}[h!]
\centering
\qquad \qquad
\includegraphics[width=2.05\columnwidth]{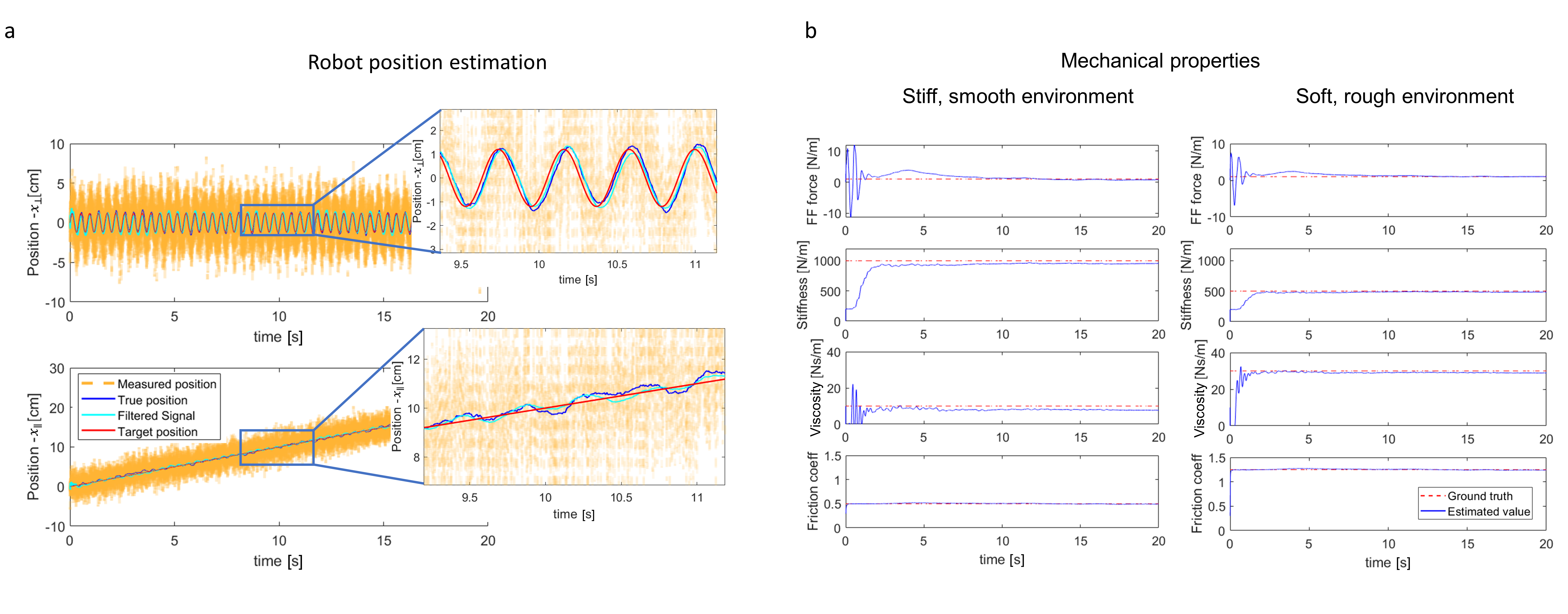}
\caption{Estimator in simulation. (a) Filtering of the robot position in the normal (top) and tangential (bottom) directions. (b) Identified mechanical properties of the two environments. From top to bottom: feedforward force, stiffness, viscosity and friction coefficient.}
\label{fig:sim_resullts}
\end{figure*}
\subsubsection{Robot's states estimation} 
The robot's states $\xi$ and friction parameter $\mu$ will be estimated together by using the nonlinear stochastic state-space model (\ref{eq:state_space_aug}), with the linearization
\begin{equation}
\begin{split}
\xi_{k+1} &= \, A_k\, \xi_{k} + B_k \, u_{k} + \omega_k \\
\eta_{k} &= \, C_{k} \, \xi_{k} + \nu_k
\end{split}
\label{eq:linear_state_space1}
\end{equation}
\begin{equation}
\begin{aligned}
A_k & = \left.\frac{\partial f(\xi,u,\theta)}{\partial \xi}\right\vert_{(\hat{\xi}_k, \hat{\theta}_k,u_k)} =\begin{bmatrix}
1 & \triangle & 0 & 0 & 0 \\
0 & 1 & 0 & 0 & 0 \\
0 & 0 & 1 & \triangle & 0 \\
0 & 0 & 0 & 1 & \hat{F}_{\perp k}/m_\parallel \\
0 & 0 & 0 & 0 & 1 \\
\end{bmatrix},\\
% \label{eq:transistion_matrix_state1}
B_k & =\left.\frac{\partial f(\xi,u,\theta)}{\partial u} \right\vert_{(\hat{\xi}_k,\hat{\theta}_k,u_k)} =\begin{bmatrix}
0 & 0 \\
0 & \triangle/m_\perp \\
0 & 0  \\
\triangle/m_\parallel & 0  \\
0 & 0 \\
\end{bmatrix},\\
C_k & = \left.\frac{\partial h(\xi)}{\partial \xi}\right\vert_{(\hat{\xi}_k,\hat{\theta}_k,u_k)} = \begin{bmatrix}
 1 & 0 & 0 & 0 & 0 \\
 0 & 0 & 1 & 0 & 0
\end{bmatrix},
% \label{eq:transistion_matrix_state1}
\end{aligned}
\end{equation}
where $m_\perp$ and $m_\parallel$ are the mass in the normal and tangential directions, and $\triangle$ is the integration time step. \(\hat{F}_\perp\) is the estimated normal force from the environment model (\ref{eq:interaction_force}). The Kalman Filter to estimate $\xi$ is then designed as
\begin{equation}
\hat{\xi}_{k+1} = \, \hat{\xi}_{k+1}^-+ K_{\xi, k+1}(\eta_{k}-C \hat{\xi}_{k+1}^-) \\
\label{eq:state_estimate_3}
\end{equation}
where $\hat{\xi}_{k+1}^- = f(\hat{\xi}_{k},u_{k},\hat{\theta}_{k})$ is the predicted states obtained by using the last estimated states and $K_{\xi, k+1}$ is the filter gain for state estimation. 

\subsubsection{Viscoelasticity parameters estimation}
The stiffness parameter $\theta$ can be estimated by using the measured normal force and interaction force model (\ref{eq:interaction_force}), and the estimated robot's state $\hat{\xi}$. %The goal is to update the viscoelasticity parameters to minimize the prediction error between force prediction and measurement. 
The EKF is used to estimate viscoelasticity parameters by considering the following state-space model:   
\begin{equation}
\begin{split}
\theta_{k+1} &= \theta_k + \omega_k \\
\eta_{\theta,k} &= h_\theta\left(\xi_{k},\theta_{k}\right) + \nu_k \,.
\end{split}
\label{eq:state_space2}
\end{equation}
The observer for the estimation of viscoelasticity parameters is given by:
\begin{equation}
\hat{\theta}_{k+1} =\hat{\theta}_{k}+K_{\theta,k+1}(\eta_{\theta,k}-C_{\theta,k}\hat{\theta}_{k})
\label{eq:estimate_W}
\end{equation}
where $K_{\theta,k+1}$ is the Kalman filter gain for parameter estimation, $\eta_{\theta,k}$ is the measured normal force, and the output matrix is
\begin{equation}
C_{\theta,k} = \left.\frac{\partial h_\theta^T \!\left(\xi_{k},\theta\right)}{\partial \theta} \right\vert_{(\hat{\xi}_k,\hat{\theta}_k)}  =\begin{bmatrix}
 1 & \hat{x}_{\perp k} & \dot{\hat{x}}_{\perp k} \\
\end{bmatrix}.
\label{eq:C_w_space}
\end{equation}\

\subsection{Interaction control}
\label{controller}
To enable the robot to smoothly track a predefined trajectory $r = [x_{\perp r},x_{\parallel r}]^T$ during the interaction with the environment,  an interaction controller using the estimated mechanical parameters is defined through
\begin{equation}
u_k = \iota_k + \phi_k \, .
\end{equation}
The feedforward component $\iota$ compensates for the interaction force using the predictive model (\ref{eq:interaction_force}). It is updated recursively with the estimated mechanical properties according to 
\begin{equation}
\iota_k = -\left[\!\!\begin{array}{c} \hat{F}_{\perp k} \\ \hat{F}_{\parallel k} \end{array} \!\!\right].
\end{equation}
The feedback component to track the target trajectory is defined as
\begin{equation}
\phi_k = - K_P \, e_k - K_D \, \dot{e}_k
\end{equation}
with the error $e_k = x_k - r_k$ and control gains $K_P$ and $K_D$. To avoid overloading while in contact with a stiff surface, the control input is saturated: $\tilde{u}_{k} = sat_M(u_{k})$ with 
\begin{equation}
\begin{aligned}
sat(s) = \left\{\begin{matrix} 
& s \quad & \!\!\!\!\!\!\!\!\!\! |s|\leq M \\
& -M \,\,\,\quad & \,\, s<-M<0 \\
& M \quad & s>M>0 \, . 
\end{matrix} \right. 
\end{aligned}
\end{equation} 

%%%%%%%%%%%%%%%%%%%%%%%%%%%%%%%%%%%%%%%%%%%%
\section{Validation of mechanical properties' estimation}

\subsection{Simulation}
We first test the designed estimator through simulating a robot interacting with different environments. A desired trajectory was designed as
\begin{equation}
r =\begin{bmatrix}
 x_{\perp r}  \\
 x_{\parallel r}  
\end{bmatrix}
=\begin{bmatrix}
 0.012\sin(15t) \\
 0.01t
\end{bmatrix} m\,, \quad t \in [0,20]\,s. 
\label{eq:desired trajectories}
\end{equation}
A sinusoidal movement was used in the normal direction that satisfies the persistent excitation condition \cite{Ioannou2016} thus ensuring that the estimator has suitable information to identify the viscoelastic parameters. The amplitude and frequency were adjusted according to the allowed surface deformation. In the tangential direction, sliding with constant speed was used to yield a homogeneous lateral contact.  

Two objects with different mechanical properties were considered: a stiff-and-smooth surface with \{$F_0$\,=\,1\,N, k\,=\,1000\,N/m, d\,=\,10\,Ns/m, $\mu$\,=\,0.5\} and a soft-and-rough surface with \{$F_0$\,=\,1\,N, k\,=\,500\,N/m, d\,=\,30\,Ns/m, $\mu$\,=\,1.25\}. The control and estimation parameters used in the simulations were \{$K_P$\,=\,1000\,kg/s$^2$, $K_D$\,=\,200\,kg/s, $P_{\xi,0}$\,=\,10$\,I_5$, $P_{\theta,0}$\,=\,5\,$I_3$, $\triangle$\,=\, 0.001\,s\} where $I_5$, $I_3$ are identity matrices, sensory noise covariance R\,=\,4\,10$^{-4}$, process noise covariance Q\,=\,2.5\,10$^{-3} I_5$.

Fig.\,\ref{fig:sim_resullts}a shows that the estimator identifies position and velocity to a value close to the real one even with large measurement noise. The estimated kinematic values were then fed back to the controller. As a result, the robot could track the target positions during the interaction in both environments. The estimated mechanical properties of the objects are shown in Fig.\,\ref{fig:sim_resullts}b. The estimation was stabilized after an approximately 2\,s settling period to a value close to the ground truth (shown in Table \ref{table:RMSE_sim}) for all mechanical properties in both stiff and compliant environments. This shows that the mechanical properties can be estimated together with the robot’s states for different objects. Note that the coefficient of restitution was not estimated as it is not involved in the continuous interaction and computed directly from (\ref{eq:CoR}).
% table 
\begin{table}[h]
\caption{Average value of mechanical parameters in the interval [10,20]\,s obtained by the estimator in simulation.}
\label{table:RMSE_sim}
\begin{center}
\begin{tabular}{| l | c | c | c | }
\hline
Object surface & \multicolumn{3}{ c |}{Estimated values (ground truth values)}  \\ 
\cline{2-4}
& \(\hat{\kappa}\) [N/m] & \(\hat{d}\) [Ns/m] & \(\hat{\mu}\) \\
\hline
stiff and smooth & 953.19 (1000)  & 7.97 (10) & 0.49 (0.5) \\ \hline
soft and rough  & 487.20 (500)  & 29.04 (30) & 1.24 (1.25)\\ \hline

%stiffness and smooth & 0.2459 & \(46.6546^1\)  & 2.288 & 0.0086 \\ \hline
%soft and rough & 0.0125 & \(14.3933^2\)  & 1.0961 & 0.0098\\ \hline

%stiff and smooth & \(22.02^*\) & 4.67  & \(20.74^*\) & 1.31 \\ \hline
%soft and rough & 5.49 &  2.73  &  3.2784 & 0.6109\\ \hline

\end{tabular}
%\begin{tablenotes}
 %   \item[] $^1$ 4.67 \(\%\) RMSPE 
  %  \item[] $^2$ 2.73 \(\%\) RMSPE
  %\end{tablenotes}
\end{center}

\end{table}

%figure
\begin{figure}[!bt]
\centering
\qquad \qquad
\includegraphics[width=0.98\columnwidth]{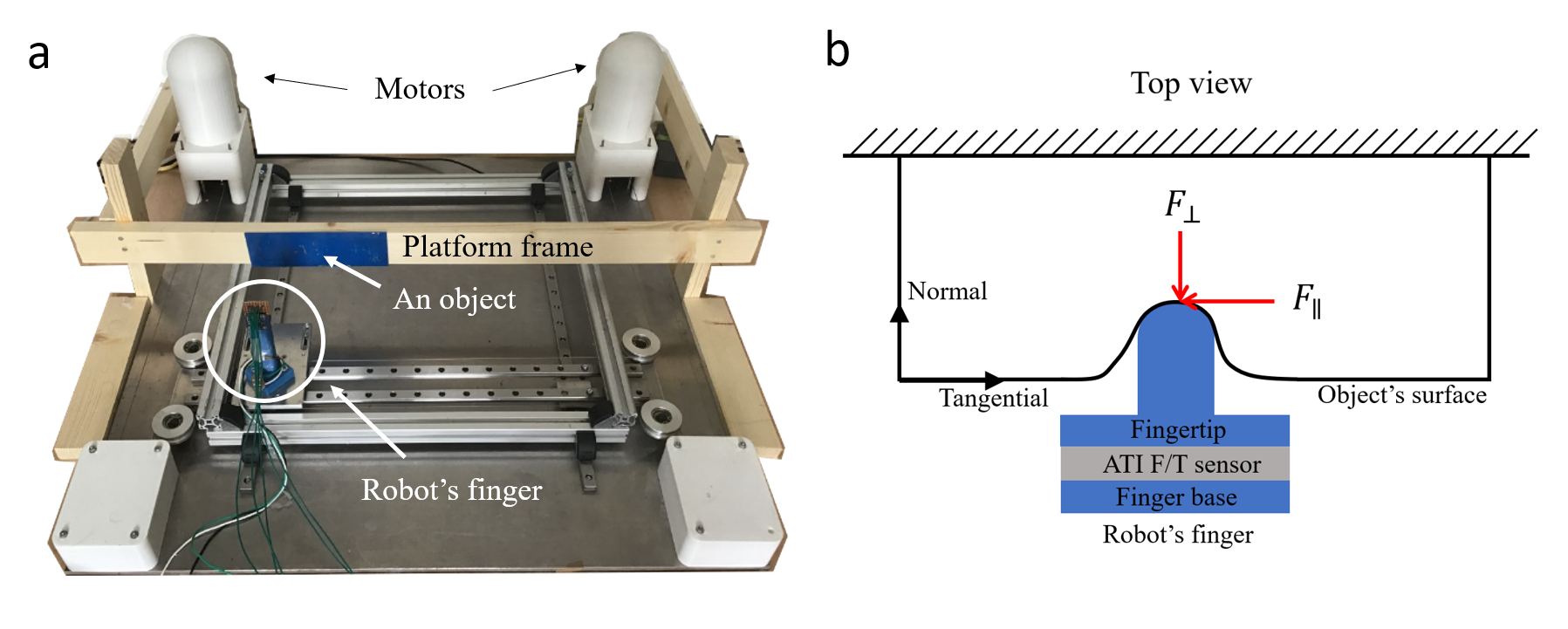}
\includegraphics[width=1\columnwidth]{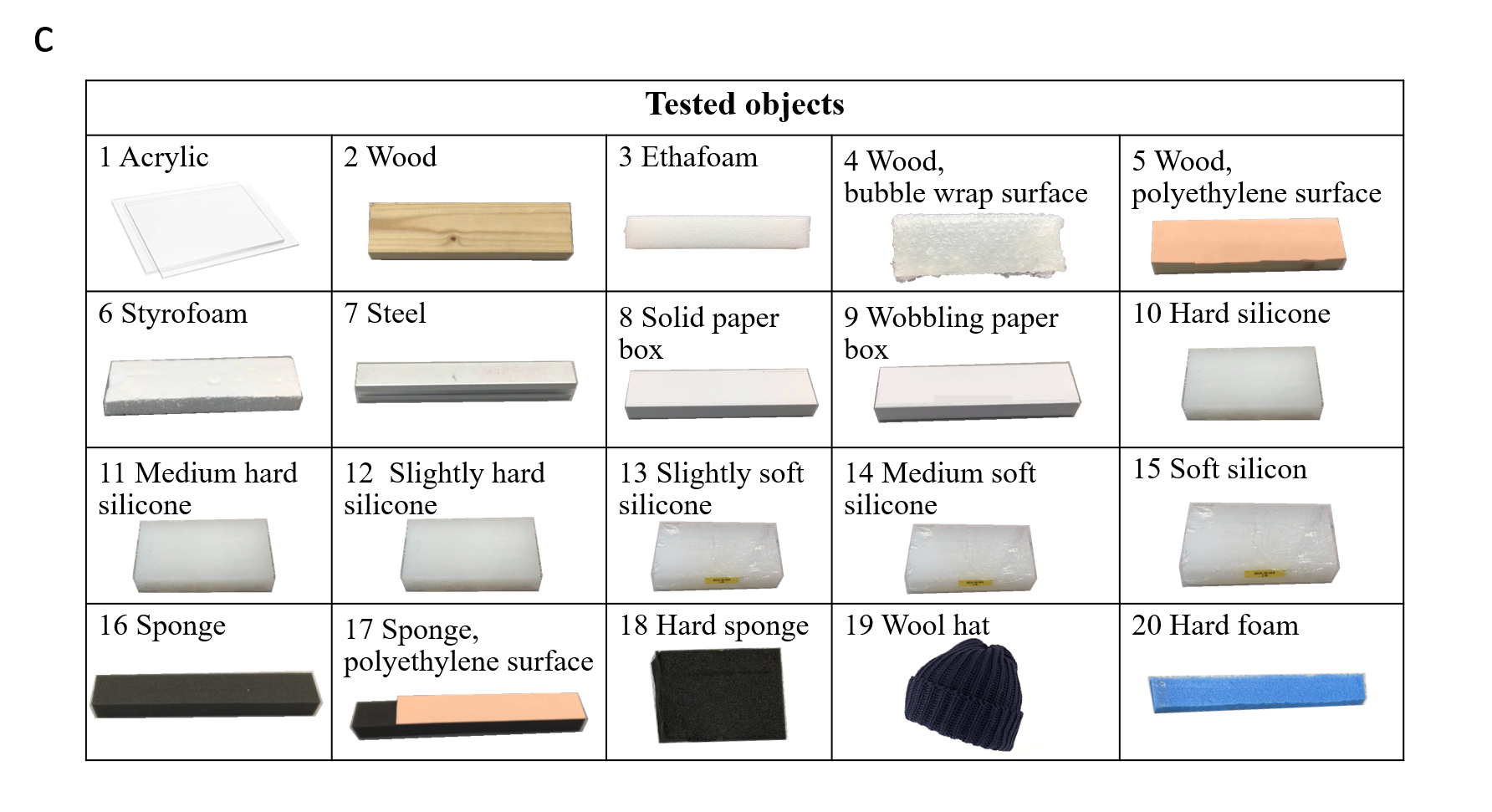}
\caption{Experimental setup. (a) HMan robot with a sensorized finger and an object to examine. A wooden platform frame is used to attach various objects for the robot to explore. The finger is thus driven by two motors in normal and tangential direction to the object's surface, where the force sensor is facing it. (b) Diagram of robot's finger interacting with an object's surface. (c) Objects used in the experiment.}
% (1) soft silicone  (2) sponge (3) sponge with polyethylene foam surface (4) hard silicone (5) wobbling box of paper (6) solid box of paper (7)styrofoam (8) wood with polyethylene foam surface (9) wood and (10) steel.
\label{fig:H-man}
\end{figure}

\subsection{Experimental validation}
\label{experiment_setup}
To validate the designed estimator with real objects and collect data for object's recognition, the HMan robot \cite{Campolo2014} shown in Fig.\,\ref{fig:H-man}a used its finger to interact with objects while estimating their mechanical properties. A six-axis force sensor (SI-25-0.25; ATI Industrial Automation) was mounted between the tip and base of the robot's finger shown in Fig.\ref{fig:H-man}b to measure the interaction forces. Note that no tangential force was required for real-time mechanical parameter estimation.
%It is noted that only the normal force was used for real-time mechanical parameter estimation.  

The robot interacted with the 20 objects shown in Fig.\,\ref{fig:H-man}c. The estimator was implemented on the robot to identify the mechanical properties of tested objects. Three actions were implemented to haptically explore the objects: 
\begin{itemize}
\item \textit{Tapping}: The robot made a first contact with the objects in the normal direction to estimate surface's elasticity. 
\item \textit{Indentation}: The robot moved its finger on the objects over the normal direction with the desired trajectory $x_{\perp r}=0.01\sin(8\tau)+0.01\,m$, $\tau \in (0,20]\,s$ to estimate the surface impedance. 
\item \textit{Sliding}: The robot slid its finger in the tangential direction along with object's surface at 0.04\,m/s while applying a 4\,N large force in the normal direction.
\end{itemize}

\begin{figure}[tb]
\centering
\qquad \qquad
\includegraphics[width=1\columnwidth]{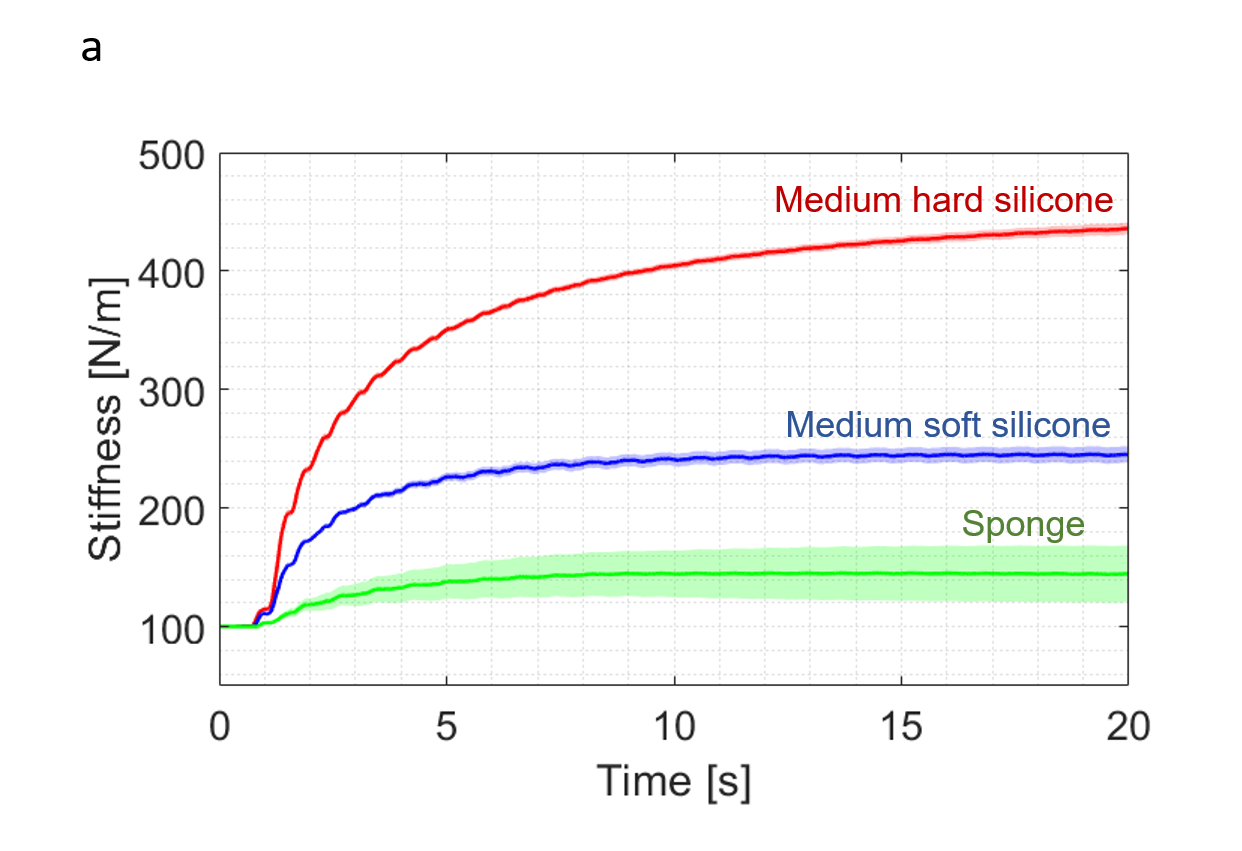}
\includegraphics[width=1\columnwidth]{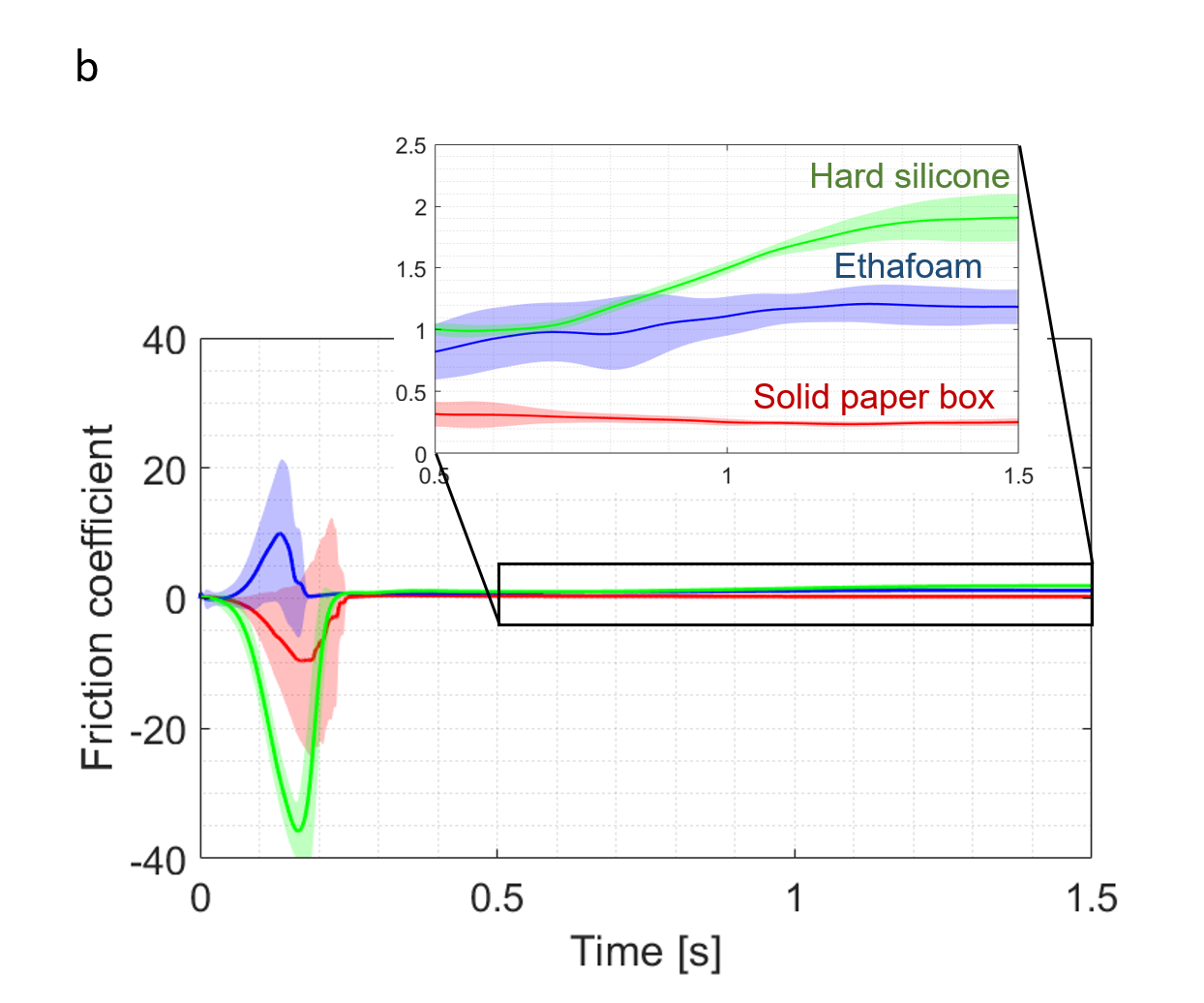}
\caption{Example of estimated mechanical properties values as a function of time. (a) Stiffness values obtained from three soft objects. (b) Friction coefficient values obtained from three hard objects.}
\label{fig:Experiment_results}
\end{figure}

The estimation was validated through 25 trials for each pair of action and object.  The resulting estimated coefficient of restitution are listed in Table \ref{table:tappeak} for four representative objects, while Fig.\,\ref{fig:Experiment_results} shows the estimated stiffness and friction coefficient estimations for some example trials. These results show that the estimations converged and that they resulted in unique values for different objects.

%table
\begin{table}[h]
\caption{Estimated coefficient of restitution}
\label{table:tappeak}
\begin{center}
\begin{tabular}{| l | c | }
\hline
Objects & $\widehat{\psi}$ \\
\hline
\hline

Sponge, polyethylene surface & 0.118 $\pm$\,0.0026 \\ \hline
Wool hat & 0.441 $\pm$\, 0.0558 \\ \hline
Acrylic & 0.299 $\pm$\,0.0258 \\ \hline
Steel & 0.632 $\pm$\,0.0401 \\ \hline

\end{tabular}
\end{center}

\end{table}

%figure

%figure
%\begin{figure}[tb]
%\centering
%\qquad \qquad
%\includegraphics[width=0.95\columnwidth]{Figures/stiffness_soft_3.PNG}
%\includegraphics[width=0.95\columnwidth]{Figures/friction_hard_3.PNG}
%\caption[]{The estimated mechanical value of example tested objects (a) stiffness from soft object group (b) friction coefficient from hard object group.}
%\label{fig:Experiment_results_2}
%\end{figure}

%%%%%%%%%%%%%%%%%%%%%%%%%%%%%%
\section{Objects' recognition}
\label{object_reg}
Object recognition was performed using the experimental data of Section \ref{experiment_setup}. The estimated coefficient of restitution was directly used as a feature. The mean values of the estimated stiffness, viscosity and friction coefficient from the last $2\,s$ of interaction were used to extract the steady-state values for additional features.

To compare the object recognition enabled by the mechanical property features with that of previous methods used in the literature, 35 such statistical features were extracted from the raw force data. Considered were the mean and maximum values, and the standard deviation from the interaction force in each direction as well as from their magnitude were extracted. In addition, a value of the normal interaction force from the first contacts was used as another feature (referred to as the ``tap peak``). The frequency spectrum of the force in both directions was obtained using a fast-Fourier transformation (FFT), which was averaged into four frequency bands: [0,35], [36,65], [66,100] and [101,500]\,Hz, where these intervals were identified in a preliminary data examination to characterise the interaction. The mean values of vibration amplitude for the frequency bands were also used as features.

To avoid overfitting, these statistical features were ranked using a feature selection method. The Chi-square test was applied since it is commonly used to evaluate features by testing their independence with class label. Finally, five feature sets were also formed to recognize objects, based on the mechanical properties features, statistical features and empirical mechanical properties features as shown in Table \ref{table:sum_set_features}.

\begin{table}[h]
\caption{Features' sets used to recognize objects.}
\label{table:sum_set_features}
\begin{center}
\begin{tabular}{|l|l|lll}
\cline{1-2}
\multicolumn{1}{|c|}{Denomination} & \multicolumn{1}{c|}{Features}   \\ \cline{1-2}
MP: Mechanical properties              & \begin{tabular}[c]{@{}l@{}}Estimated stiffness,\\ viscosity, friction coefficient,\\ coefficient of restitution.\end{tabular} \\ \cline{1-2}
SF: Statistical features            &  \begin{tabular}[c]{@{}l@{}} 35 statistical features. \end{tabular}                     \\ \cline{1-2}
\begin{tabular}[c]{@{}l@{}} CSSF: Chi-square test \\ statistical features \end{tabular}                 &\begin{tabular}[c]{@{}l@{}} The first 4 statistical features \\ ranked by Chi-square test; tap peak, \\ mean $F_{\parallel}$ obtained by sliding, \\ std magnitude obtained by pressing, \\ mean \(F_{\perp}\) obtained by pressing. \end{tabular}  \\ \cline{1-2}
\begin{tabular}[c]{@{}l@{}} EMP1: Empirical mechanical \\ properties feature 1 \end{tabular}                & \begin{tabular}[c]{@{}l@{}}1 feature for stiffness, \\ 3 features for surface's texture, \cite{Kaboli2019}.\end{tabular}                \\ \cline{1-2}
\begin{tabular}[c]{@{}l@{}} EMP2: Empirical mechanical \\ properties feature 2 \end{tabular}             & \begin{tabular}[c]{@{}l@{}}2 features for compliance\\ and texture, \cite{Xu2013}.  \\ 2 features for surface's roughness\\ and fineness, \cite{Fishel2012}.\end{tabular}                     \\ \cline{1-2}
\end{tabular}
\end{center}
\end{table}

These feature's sets were used to evaluate their performances in object recognition using supervised and unsupervised learning methods as described in Table \ref{table:sum_exp}. The Naive Bayes classifier was selected as it exhibited superior performance than other classifiers for supervised learning. Gaussian mixture models (GMMs) were used to investigate clustering with unknown labels. 
These clustering results were evaluated by comparing them with the known labels using normalised mutual information defined as: 
\begin{equation}
\text{NMI} =\frac{2\,MI(C;L)}{H(C)+H(L)} \\
\label{eq:NMI}
\end{equation}
where $MI(C;L)$ is the mutual information between a set of clustering results $C=\{c_1, c_2,...c_N\}$ and known labels $L=\{l_1, l_2,...l_N\}$:
\begin{equation}
MI(C;L) =\sum_{i}\, \sum_{j}\, p(c_i \cap l_j)\,\log \,\frac{p(c_i \cap l_j)}{p(c_i)\cdot p(l_j)}\ \ \\
\label{eq:NMI}
\end{equation}
and \(H(\cdot)\) is an entropy: 
\begin{equation}
H(X)= -\sum_{\substack{x \in X}}\,p(x)\, \log\,p(x) \,.
\label{eq:entropy}
\end{equation}
NMI evaluates how random the generated clusters are with respect to the known labels in a range of [0,1], where 1 means the clusters are perfectly generated according to the known labels and 0 that they are generated randomly.

\begin{table}[h]
\caption{Learning method, feature selection algorithms and dataset used for classification. The abbreviations are defined in Table \ref{table:sum_set_features}.}
\label{table:sum_exp}
\begin{center}
\begin{tabular}{|l|l|l|l}
\cline{1-3}
\multicolumn{1}{|c|}{\begin{tabular}[c]{@{}c@{}}Method\end{tabular}}  &
  \multicolumn{1}{c|}{Algorithm} &
  \multicolumn{1}{c|}{Dataset} &
   \\ \cline{1-3}
Supervised & Naive Bayes  & \begin{tabular}[c]{@{}l@{}}MP\end{tabular} &  \\ \cline{1-3}   
   
Supervised & Naive Bayes  & \begin{tabular}[c]{@{}l@{}}MP\\ SF\\ CSSF\\ EMP1\\ EMP2\end{tabular} &  \\ \cline{1-3}
Unsupervised    & GMMs         & as in supervised                                                                  &  \\ \cline{1-3}
\end{tabular}
\end{center}
\end{table}

%figure
\begin{figure*}[h!]
\centering
\qquad \qquad
\includegraphics[width=1.6\columnwidth]{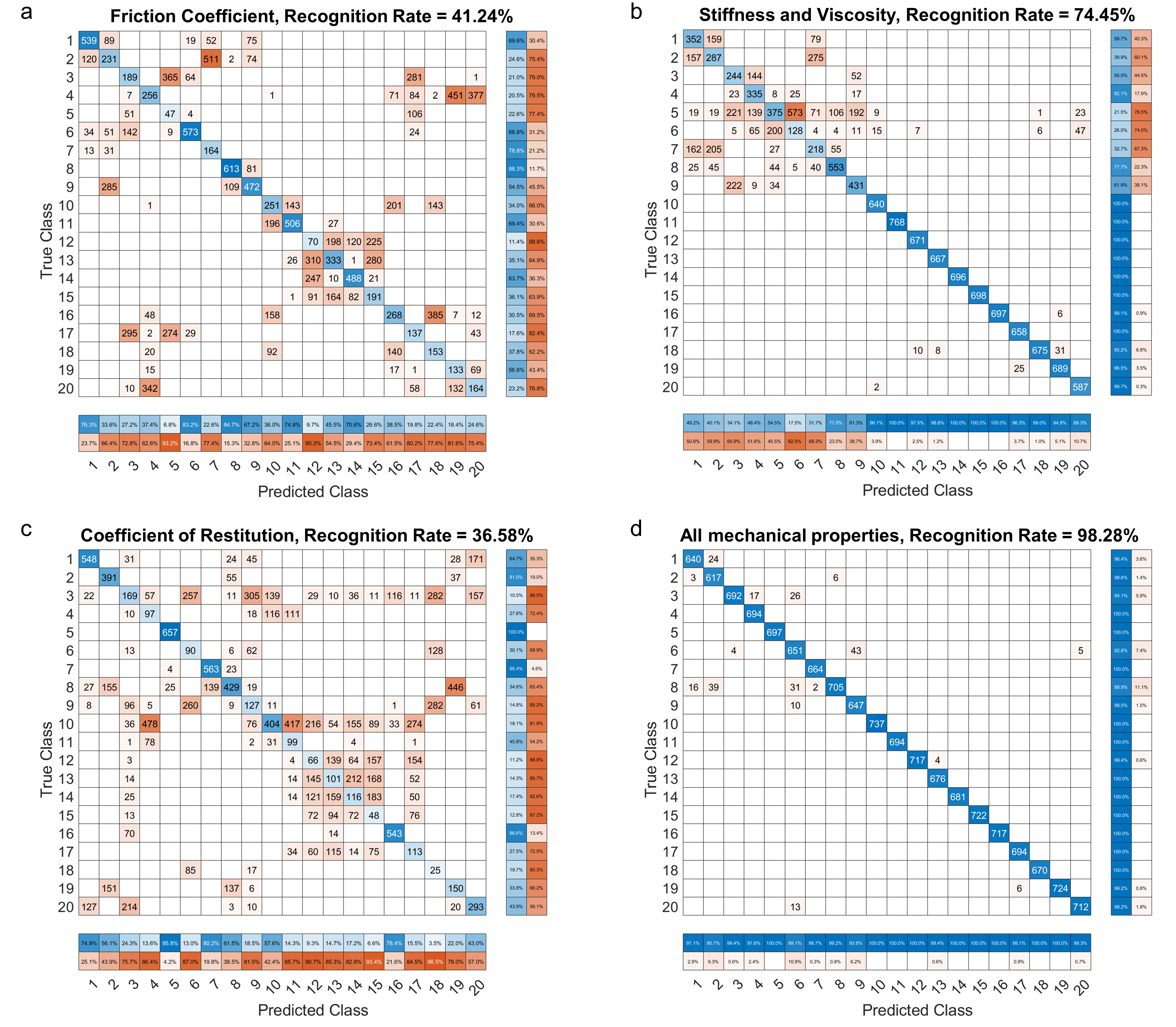}
\caption[]{Confusion matrix obtained  by using (a) friction coefficient (b) stiffness and viscosity (c) coefficient of restitution (d) all mechanical properties as features. A colour blue corresponds to correct classified objects while a colour red corresponds to an incorrect classified objects.}
\label{fig:confusion_matrix}
\end{figure*}

\subsection{Classification with mechanical properties}

To understand how each of the estimated mechanical properties impact objects' classification, an estimation using a single feature was first performed, then gradually other features were added to feed the classifier. The classification was evaluated through a four-fold cross-validation using a 3:1 train:test ratio with 100 repetitions.

Fig.\,\ref{fig:confusion_matrix} shows the object recognition confusion matrix results using the friction coefficients (a), stiffness and viscosity (b), coefficient of restitution (c) and all estimated mechanical properties (d). It can be seen that by using only friction or the coefficient of restitution, the classifier cannot recognise all the objects, resulting in a recognition rate lower than 50\%. These features only recognise a group of hard objects (classes 1-10) better than soft objects (classes 11-20) as shown in Figs.\,\ref{fig:confusion_matrix}a,c. Using stiffness and viscosity increased the recognition rate to 74.45\%, but could not differentiate hard objects (Fig.\,\ref{fig:confusion_matrix}b). 

By using all four mechanical properties in the classifier, the recognition rate increased to 98.18\% (Fig.\,\ref{fig:confusion_matrix}d). The resulting confusion matrix exhibits almost perfect recognition with a rate over 90\% for each object. There still is some confusion between pairs of object class, but for each object the misclassification rate is lower than 0.05\,\%, which can be considered as negligible. These results demonstrate the advantages provided by using the combination of different mechanical properties in order to classify various objects.

%%%%%%%%%%%%%%%%%%%%%%%%%%%%%%
\subsection{Objects' classification with mechanical properties vs. statistical features}
\label{Classification with various datasets}
To examine the role of using the estimated mechanical properties for object classification compared to other sets of features, the classifier was used to find the recognition rates from the five sets of features described in Table III: mechanical property features (MP), statistical features (SF and CSSF) and empirical mechanical properties features (EMP1 and EMP2). These object classifications were evaluated by a four-fold cross-validation using 100 repetitions. 

Fig.\,\ref{fig:acc_results}a shows that using mechanical properties as features resulted in a recognition rate of 98.18\,$\pm$\,0.424\,\%. On the other hand, the statistical features with and without feature selection resulted in a recognition rate of 92.2\,$\pm$\,0.60\,\% and  89.7\,$\pm$\,3.20\,\% respectively. Lastly, features used in \cite{Kaboli2019} reflecting on EMP1 provides 77.5\,$\pm$\,5.07\,\% and features used in \cite{Fishel2012} and \cite{Xu2013} reflecting on EMP2 yields 82.9\,$\pm$\,0.91\,\%. These results show that mechanical properties provided the highest recognition rate while using a lower number of features and without needing tangential force sensing.

\subsection{Objects' clustering using mechanical properties or statistical features}
\label{clustering}
To study the benefit of using the coefficient of restitution  stiffness, viscosity and friction coefficient together in an unsupervised learning method, GMMs clustering was used to perform a clustering task with the same five sets of feature as in section \ref{Classification with various datasets}. We assumed that each cluster had its own diagonal covariance matrix and the number of clusters is set to 20, i.e. the number of tested objects. This clustering task was done and evaluated by NMI for 40 repetitions for each set of features.

\begin{figure}[!bt]
\centering
\qquad \qquad
\includegraphics[width=1\columnwidth]{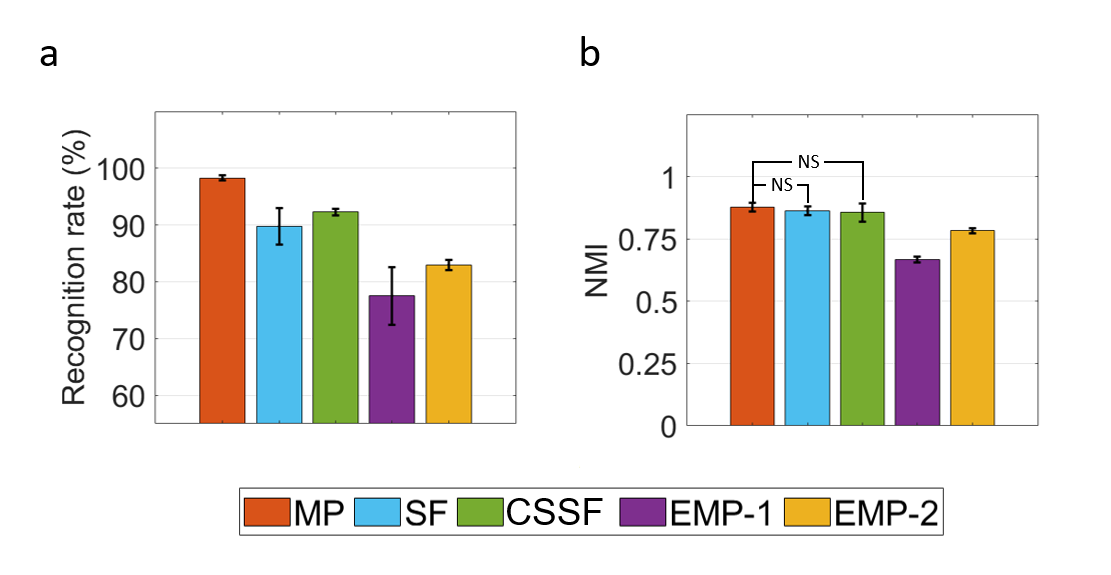}
\caption{Comparison of classification (a) and clustering using normalized mutual information (NMI) (b) with different feature sets described in Table III.}
\label{fig:acc_results}
\end{figure}

The evaluation of clustering results using NMI is shown in Fig.\,\ref{fig:acc_results}b. Using mechanical features as input data in a clustering task gave NMI values of 0.851\,$\pm$\,0.03  which is similar to what the SF and CSSF provided at 0.863\,$\pm$\,0.016 and 0.856\,$\pm$\,0.018 respectively (\emph{p}$>$0.05). However, the NMI results obtained using the MP were found to be significantly higher than the NMI results obtained using EMP1 and EMP2 (\emph{p}$<$0.05). These results suggest that using four mechanical properties could provide the same results as with 35 statistical features. In addition, it also could outperform the other features representing empirical mechanical properties used in \cite{Kaboli2019}, \cite{Fishel2012} and \cite{Xu2013}.

%figure

%%%%%%%%%%%%%%%%%%%%%%%%%%%%%%%%%%%
\section{Discussion}
This paper introduced an object recognition framework based on the estimation of mechanical properties with a dual extended Kalman filter. This online identification extends \cite{Li2018} and stably estimates the coefficient of restitution, stiffness, viscosity and friction parameters. The viability of this method was demonstrated in simulations and experiments. %Using the intrinsic representation with the selected material properties, the recognition rate did not depend much on the algorithm used for classification.

The classification performance was evaluated with 20 real-world objects. Using the four representative mechanical parameters, a recognition rate of 98.18\,$\pm$\,0.424\,\% could be achieved using supervised learning, and clustering exhibited a normalized mutual information of 0.851\,$\pm$\,0.03. Using only four mechanical properties resulted in a better classification and similar clustering as with 35 statistical features, suggesting that mechanical features entail a more compact and accurate representation than statistical features.

Note that all of the coefficient of restitution, viscoelastic and friction parameters were required to distinguish objects. In particular, including the coefficient of restitution largely improved the recognition rate in comparison to when using only viscoelastic parameters. For example, using stiffness could not distinguish steel from wood as both are hard materials, but they had different impact properties as measured by the coefficient of restitution.

The intrinsic mechanical properties identified in our scheme provided better and more consistent results than the empirical mechanical properties used in previous works \cite{Kaboli2019, Xu2013, Fishel2012}. This illustrates the limitations of using empirical features to recognize objects, which may depend on the specific action used. For instance, the surface texture measure of \cite{Xu2013} was defined as a variance of the interaction force in the normal direction during a robot's finger sliding on object's surfaces, which may depend on the object's pose and robot's interaction and lead to inconsistent estimation results.

In summary, this work emphasized the role of mechanical properties in haptic exploration, and how they can be used to reliably recognise different objects. The results demonstrated the superiority of the mechanical properties-based object recognition, yielding more reliable recognition than empirical properties and requiring far less features than based on statistical features. Moreover, using this intrinsic object representation makes the framework flexible to using various classification algorithms. While the presented system could successfully recognize objects during haptic exploration, considering the weight and inertial parameters would enable extending our framework from haptic exploration to enabling full object manipulation with transport.

\printbibliography[]
% \bibliographystyle{unsrt}    
% \bibliography{references} 

\appendices
\section{Full results for classification with mechanical properties}
\label{FirstAppendix}
\begin{figure}[H]
\centering
\includegraphics[width=0.48\textwidth]{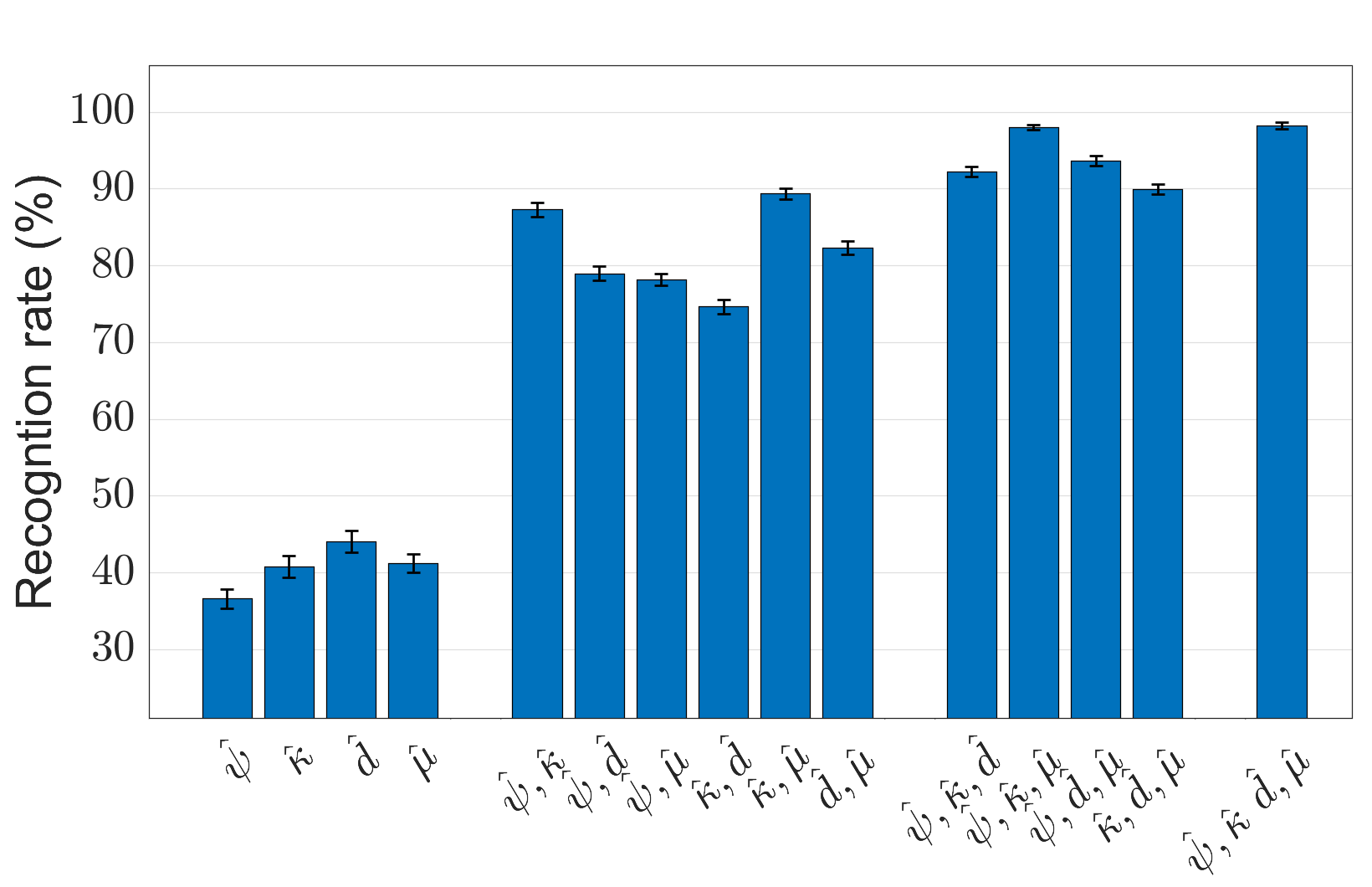}
\caption{\label{fig:mech_results}Classification results with different combinations of the mechanical features. $\hat{\psi}$ is the coefficient of restitution, $\hat{\kappa}$ is stiffness, $\hat{d}$ is viscosity and $\hat{\mu}$ the friction coefficient.}
\end{figure}

Fig.\,\ref{fig:mech_results} shows the classification results from all combinations of mechanical properties used as features in the Naive Bayes classifier. Starting from a single feature until four features, the recognition rate increased as the number of mechanical properties increase. The highest value can be achieved by using all four estimated mechanical proprieties. 

\end{document}